\documentclass[runningheads]{llncs}
\usepackage{eccv}
\usepackage{eccvabbrv}
\usepackage{graphicx}
\usepackage{booktabs}
\usepackage[accsupp]{axessibility}
\usepackage{bm}
\usepackage{multirow}
\usepackage{adjustbox}
\usepackage{makecell}
\usepackage{xcolor}
\usepackage{array}
\usepackage{colortbl}
\usepackage{url}
\usepackage{tabularx}
\usepackage{wrapfig}
\usepackage{threeparttable}
\usepackage{algorithm}
\usepackage{algpseudocode}
\usepackage[most]{tcolorbox}
\usepackage{cuted}
\usepackage{hyperref}
\usepackage{orcidlink}
\usepackage{tablefootnote}

\newcommand{\para}[1]{
  \vspace{1.2mm}
  \noindent\textbf{#1. }
}

\newcommand{\rqone}[0]{
To what extent does MLLM-as-a-Judge exhibit self-preference bias?
}
\newcommand{\rqtwo}[0]{
To what extent does cross-model preference bias appear in MLLM-as-a-Judge?
}
\newcommand{\rqthree}[0]{
Can an ensemble of Evaluators mitigate the influence of model-specific preference bias while maintaining alignment with human judgments?
}
\newcommand{\TColorBox}[2]{
\begin{tcolorbox}[enhanced, breakable,
  colback=blue!5!white,
  colframe=blue!75!black,
  fonttitle=\bfseries,
  title=#1,
  boxrule=0.5pt,
  sharp corners,
  left=2mm, right=2mm, top=1mm, bottom=1mm,
  before upper={\let\\\par},
]
#2
\end{tcolorbox}
}

\begin{document}
\title{
\hspace{-1mm}MLLM-as-a-Judge Exhibits Model Preference Bias\hspace{-1mm}
}
\newcommand*\samethanks[1][\value{footnote}]{\footnotemark[#1]}

\author{
Shuitsu Koyama\thanks{Equal contribution}
\and
Yuiga Wada\samethanks
\and
Daichi Yashima
\and
Komei Sugiura
}
\authorrunning{S.~Koyama et al.}
\institute{
Keio University, Japan
\\\email{\{koyamashu3, yuiga, ydaichi1207, komei.sugiura\}@keio.jp}
}

\maketitle
\begin{abstract}
Automatic evaluation using multimodal large language models (MLLMs), commonly referred to as MLLM-as-a-Judge, has been widely used to measure model performance. If such MLLM-as-a-Judge methods were biased, they could distort model comparisons and benchmark-driven scientific progress. However, it remains unclear to what extent MLLM-as-a-Judge methods favor or disfavor text generated by specific MLLMs. In this study, we propose Philautia-Eval to investigate such model-specific preference bias. Philautia-Eval quantifies the degree of the bias by disentangling preference tendencies from differences in generation quality. Using 1.29M caption-score pairs collected from 12 MLLMs, we found that representative MLLMs tend to exhibit self-preference bias. Moreover, experimental results indicate mutual preference bias within particular model families, which is potentially driven by reused connectors and overlapping instruction-tuning resources. Finally, we introduce a simple ensemble of MLLMs, \textsc{Pomms}. Our results demonstrated that \textsc{Pomms} effectively mitigated the model-specific preference bias while maintaining performance. Our project page is available at \url{https://philautia.kinsta.page/}.
\end{abstract}
\begin{figure}[!t]
    \centering
    \includegraphics[width=\linewidth]{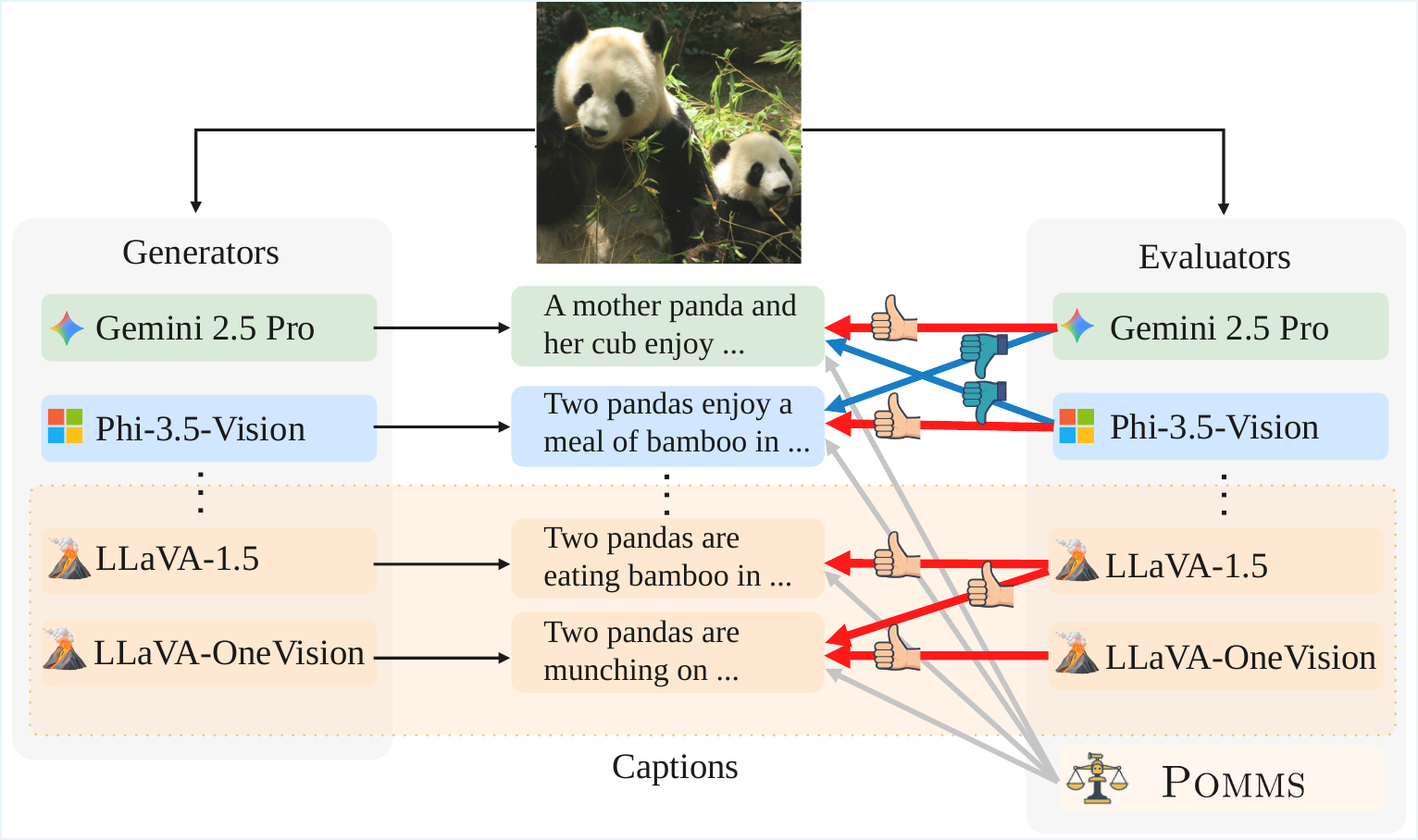}
    \caption{\textbf{Schematic of our approach for investigating model-specific preference bias in MLLM-as-a-Judge.} As described in Section~\ref{sec:exp}, each MLLM typically favors its own generations (self-preference bias), whereas LLaVA-1.5~\cite{llava-1.5} favors text generated by other models within the LLaVA family (cross-model preference bias). Our \underline{\textbf{P}}anel \underline{\textbf{o}}f \underline{\textbf{M}}LL\underline{\textbf{M}} Evaluator\underline{\textbf{s}} (\textsc{Pomms}) exhibits less model-specific preference bias.
    }
    \label{fig:eye-catch}
    \vspace{-6mm}
\end{figure}

\section{Introduction}
\label{sec:intro}
Multimodal large language models (MLLMs) are being actively researched and increasingly deployed across a wide range of social domains, and the global MLLM market is valued at approximately USD 2.69 billion \cite{market}. The essential nature of evaluation for MLLM development has led to increased use of MLLM-as-a-Judge methods, which serve as evaluators for generated outputs~\cite{tong2024gveval,lee2024fleur, mllm-as-a-judge}. If such MLLM-as-a-Judge methods were biased, they could distort benchmark-driven scientific progress in our community.

Of particular concern is model-specific preference bias~\cite{self_pref_acl24, self_pref_neurips, acl24_pride}, where a model tends to overestimate or underestimate scores for generated text by specific MLLMs. \cref{fig:eye-catch} shows a typical example of this bias, where Gemini-2.5-Pro~\cite{gemini25pro} and Phi-3.5-Vision~\cite{phi35vision} overestimate their own generated captions, while underestimating each other. With MLLM-as-a-Judge becoming a common evaluation paradigm~\cite{tong2024gveval, lee2024fleur, expert, discode}, it is important to examine the extent to which these methods exhibit such bias.

Despite such concerns, previous studies~\cite{self_pref_neurips, self_pref_acl24, self_pref_emnlp, mllm-as-a-judge} have not sufficiently quantified such bias because they did not appropriately disentangle preference tendencies from differences in generation quality. Existing approaches (e.g., \cite{self_pref_neurips}) assess the bias by merely asking an LLM to compare its own output against that of another model. However, an LLM may select text generated by a specific model simply because of its higher quality rather than any inherent preference. Consequently, model performance may be conflated with preference bias, leading to insufficient clarification of these tendencies.

In this study, we investigate model-specific preference bias in MLLM-as-a-Judge for image captioning. To this end, we introduce Philautia-Eval, which quantifies the degree of the bias by disentangling preference tendencies from differences in generation quality. Using Philautia-Eval, we analyzed 1.29M caption--score pairs collected from 12 MLLMs.

Our key findings are as follows:
\begin{itemize}
    \item  [$\bullet$]Representative MLLMs tend to exhibit self-preference bias in both reference-based and reference-free settings for image captioning.
    \item  [$\bullet$]Experimental results indicate mutual preference bias within particular model families, presumably driven by reused connectors and overlapping instruction-tuning resources.
    \item  [$\bullet$]We introduce a simple ensemble of MLLMs, \textsc{Pomms}, which effectively mitigates the influence of model-specific preference bias, while maintaining performance.
\end{itemize}

\section{Related Work}
\label{sec:related}
There have been numerous surveys on LLM-as-a-Judge methods across a wide range of tasks~\cite{llm-as-a-judge-survey, llm-as-a-judge-survey-2}, including image captioning and retrieval. In particular, Gu et al.~\cite{llm-as-a-judge-survey} provide a detailed overview of LLM-as-a-Judge with a particular focus on reliability,  while Gallegos et al.~\cite{bias_fair_survery} provide a comprehensive summary of bias in LLMs, covering definitions of bias and mitigation methods across diverse settings.

\para{LLM-as-a-Judge}
LLM-as-a-Judge and MLLM-as-a-Judge have been used for a wide range of tasks~\cite{llm-as-a-judge-survey}.
In the automatic evaluation of document summarization and dialogue response generation, several LLM-as-a-Judge methods have been proposed~\cite{geval, checkeval, gptscore}. Other studies have explored LLM-based metrics for image captioning (e.g.~\cite{chan2023clair, yao2024hifi}). For instance, CLAIR~\cite{chan2023clair} outputs the scores of the candidate captions by prompting an LLM with candidate captions and references.

Beyond text-only judges, MLLM-as-a-Judge methods~\cite{lee2024fleur, tong2024gveval, expert, discode} that incorporate images have been developed by conditioning on the images through MLLMs. For example, FLEUR~\cite{lee2024fleur} introduces score smoothing, which calibrates the raw evaluation score using token probabilities to enable fine-grained evaluations. Similarly, G-VEval~\cite{tong2024gveval} evaluates captions based on the probabilities of output tokens using chain-of-thought reasoning.

\para{Preference bias}
LLMs have been shown to exhibit many forms of bias, including social, cultural, and demographic biases~\cite{bias_llm, culture-1, culture-2, social-1, social-2}. Furthermore, LLM-as-a-Judge has been observed to exhibit preference bias (e.g., position, length, and verbosity)~\cite{judging_judge, judgelm, not_fair_eval, emnlp_langthbias, neurips_lengthbias_verbosity}. Such bias refers to a tendency for LLM-as-a-Judge to favor certain outputs independent of their underlying quality~\cite{judging_judge, judgelm}. Several studies~\cite{judging_judge, judgelm, not_fair_eval} have revealed position bias, where  the choice made by an LLM-as-a-Judge can be changed by simply reordering the candidate answers. Furthermore, Hu et al.~\cite{emnlp_langthbias} reported on length bias, in which LLM-as-a-Judge tends to favor long sentences, which can distort assessments of response quality and trustworthiness.
Finally, Zheng et al.~\cite{neurips_lengthbias_verbosity} highlighted the problem of verbosity bias, a tendency to prefer longer responses despite a lack of additional content.

\para{Self-preference bias}
LLM-as-a-Judge methods have been shown to exhibit self-preference bias~\cite{bias_sb, self_pref_neurips, self_pref_emnlp, self_pref_reason, aiaibias, self_pref_acl24, acl24_pride}. A common evaluation approach relies on relative measurements (i.e., pairwise judgments) between candidates generated by other LLMs~\cite{self_pref_neurips, self_pref_emnlp, self_pref_reason, aiaibias}. 
For example, Panickssery et al.~\cite{self_pref_neurips} examined self-preference bias by asking an LLM to evaluate whether its own generated text is superior to text generated by another model. This is problematic because the LLM may judge its own text as superior simply because it is of higher quality, making it difficult to disentangle model performance from model preference bias.

In contrast, several prior studies have examined self-preference bias using non-pairwise measurements~\cite{self_pref_acl24, acl24_pride}. Liu et al.~\cite{self_pref_acl24} employed an LLM to evaluate text generated by both itself and other LLMs and confined their evaluation scores to the $[0, 1]$ range via min-max scaling. However, this method may erroneously yield a value of 0 even when preference bias is present, because the normalized set always contains the values 0 and 1.

\para{Datasets for image captioning}
Several datasets have been proposed for evaluating image captions with human judgments~\cite{composite, flickr, polos, deneb, pearl, vela}. Standard datasets for image caption generation include MS COCO~\cite{coco} and nocaps~\cite{nocaps}.
The nocaps dataset, which covers more diverse visual concepts than MS COCO~\cite{coco,nocaps}, consists of 15,100 images from Open Images~\cite{open-image} and 166,100 human-written captions. However, neither of these datasets can be used to investigate self-preference bias because they lack either captions generated by MLLMs or MLLM-based evaluation scores~\cite{coco,nocaps,composite, flickr, polos, deneb, pearl, vela}. Polaris~\cite{polos} is a large-scale dataset that contains 131,020 human judgments of the captions generated for 13,691 images; however, it also likewise does not include evaluation scores given by MLLMs~\cite{polos}.
Therefore, we construct SelfEval-Cap, a dataset for investigating self-preference bias. This dataset contains captions generated by 12 MLLMs together with the evaluation scores given by the corresponding MLLMs.

\section{Problem Setting}
\label{sec:problemset}
\begin{figure}[t]
    \centering
    \includegraphics[width=0.9\linewidth]{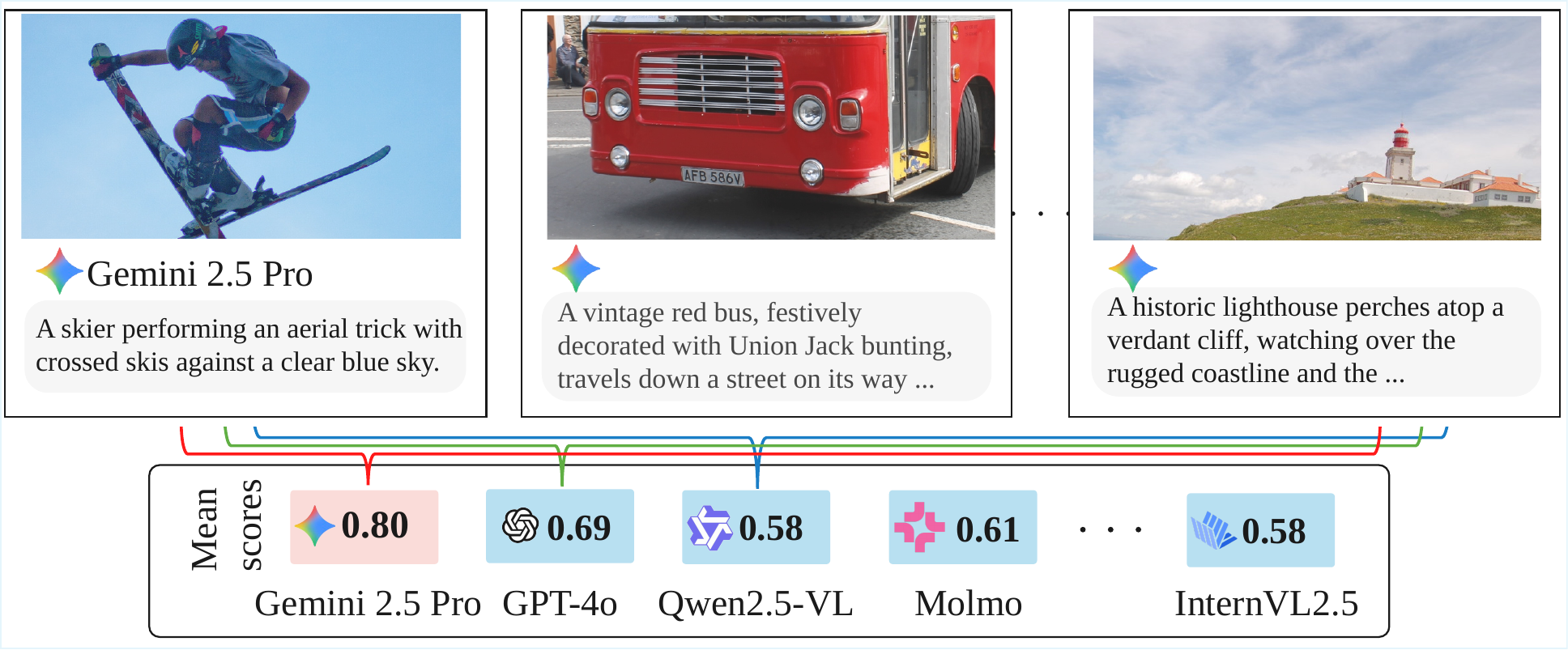}
    \vspace{-3mm}
    \caption{\textbf{An example of self-preference bias in MLLM-as-a-Judge.} The figure shows captions generated by Gemini 2.5 Pro (top) and the average scores given to these captions by Evaluators (bottom). In this example, Gemini 2.5 Pro may exhibit self-preference bias because it gives higher scores on average to its own captions than other Evaluators do.}
    \label{fig:self_pref_example}
    \vspace{-3mm}
\end{figure}

\para{Definition}
We investigate model-specific preference bias in MLLM-as-a-Judge for image captioning.
Following previous studies~\cite{self_pref_acl24, mllm-as-a-judge, self_pref_neurips, bias_sb}, we define model-specific preference bias as the tendency of an MLLM to favor specific MLLMs' generated texts over those generated by other MLLMs.
We further distinguish two cases: self-preference bias (favoring its own generations) and cross-model preference bias (favoring or disfavoring specific external models).
We refer to the MLLM used to evaluate generated captions as the Evaluator, to the MLLM used to generate captions as the Generator.
\cref{fig:self_pref_example} illustrates an example of self-preference bias in MLLMs. 
In this example, Gemini 2.5 Pro~\cite{gemini25pro} may exhibit self-preference bias because this model gives higher scores on average (0.80) to its own captions than the other Evaluators (e.g., GPT-4o~\cite{gpt4o} at 0.69).

\para{Research Questions}
MLLM-as-a-Judge could potentially exhibit model-specific preference bias for the following reason: 
As a Generator, for a given input image and a prompt, an MLLM in principle outputs a high-likelihood token sequence. 
When an Evaluator uses the same model weights, the sequence maintains high likelihood for the Evaluator.
Similarly, such a tendency may occur even if the backbone MLLMs differ.
Prior research~\cite{likelihood_acl24, bias_sb} demonstrates that LLM-as-a-Judge favors high-likelihood sequences and overestimates their scores. Therefore, we hypothesize that MLLMs inherit this behavior despite the additional visual modality, causing the Evaluator to favor specific MLLMs' generations.

While the existence of such bias is plausible, several aspects remain underexplored. First, it is unclear to what extent MLLM-as-a-Judge methods exhibit self-preference bias. 
Second, cross-model preference bias in MLLMs has not been thoroughly investigated.
Finally, methods for mitigating the bias have yet to be fully examined.

Therefore, we formulate the following three research questions:
\begin{itemize}
    \item [$\bullet$] RQ1. \rqone
    \item [$\bullet$] RQ2. \rqtwo
    \item [$\bullet$] RQ3. \rqthree
\end{itemize}

In this study, we focus on MLLM-as-a-Judge for image captioning~\cite{tong2024gveval} for the following reasons:
Image captioning is a standard task for benchmarking MLLMs, and is particularly suitable for investigating model-specific preference bias because it does not constrain the generated text to specific formats (e.g., multiple-choice answers in typical VQA).
Furthermore, we assume MLLMs that provide token-level probabilities of output tokens. This is motivated by the observation that metrics leveraging token-level probabilities (e.g., FLEUR~\cite{lee2024fleur}, G-VEval~\cite{tong2024gveval}) tend to correlate more closely with human judgments than metrics that do not (e.g., CLAIR \cite{chan2023clair}).

\section{Methodology}
\label{sec:method}
\subsection{Philautia-Eval}
We introduce Philautia-Eval, which quantifies the extent to which an Evaluator favors captions generated by a Generator. Philautia-Eval investigates model-specific preference bias by analyzing statistical tendencies of scores given to generated texts across various combinations of Generators and Evaluators. Our method can be broadly applied to the investigation of model-specific preference bias in evaluation based on generative models (e.g., LLM-as-a-Judge).

\begin{figure}[!t]
    \centering
    \includegraphics[width=\linewidth]{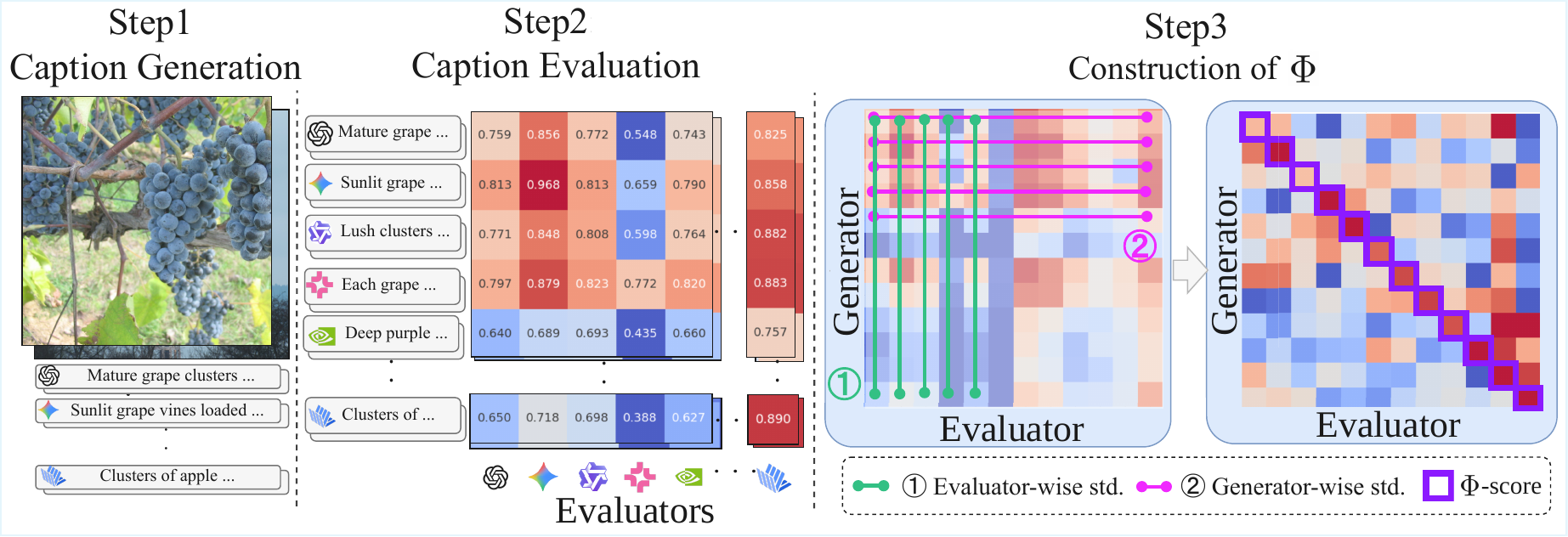}
    \caption{\textbf{Pipeline of the proposed method.} Generators generate image captions, which are then given evaluation scores by Evaluators. From these scores, we construct a matrix $\mathrm{\Phi}$, whose rows and columns correspond to the Generators and the Evaluators, respectively. We then standardize $\mathrm{\Phi}$ column-wise and subsequently row-wise to obtain $\tilde{\mathrm{\Phi}}$. The diagonal entries indicate the degree of self-preference bias, which we name the \textit{philautia score}. In the figure, ``std.’’ stands for standardization. }
    \label{fig:pipeline}
\end{figure}

\cref{fig:pipeline} and Algorithm \ref{alg:xprefeval} show the pipeline of our method and its corresponding pseudocode, respectively. As illustrated in \cref{fig:pipeline}, Philautia-Eval follows a three-stage process:

\para{Step1. Caption generation} Each Generator generates captions for the given images. For the $k$-th image ${\bm{x}_{\mathrm{img}}^{(k)}}$ ($k=1,\ldots ,N$), the $i$-th Generator $G^{(i)} \; ( i = 1, \ldots, M)$ takes  $\bm{x}_{\mathrm{img}}^{(k)}$ and a generation prompt $\bm{x}_{\mathrm{gprompt}} $ as input and outputs a caption $\hat{\bm{y}}_{g}^{(i,k)}$. Here, $N$ and $M$ denote the numbers of images and MLLMs, respectively.

\vspace{3mm}
\para{Step2. Caption evaluation} Evaluating the captions generated in Step 1 with diverse Evaluators enables preference analysis across all Evaluator–Generator pairs. 
We employ the $j$-th Evaluator $E^{(j)}$ $(j=1, \ldots,M)$ to evaluate $\{\hat{\bm{y}}_{g}^{(i,k)}\}_{i=1,k=1}^{M,N}$. Specifically, we input $\bm{x}_{\mathrm{e}}^{(i,k)}=(\bm{x}_{\mathrm{img}}^{(k)}, \bm{X}_{\mathrm{ref}}^{(k)}, \hat{\bm{y}}_{\mathrm{g}}^{(i,k)},\bm{x}_{\mathrm{eprompt}})$ into each $E^{(j)}$, where $\bm{X}_{\mathrm{ref}}^{(k)}$ and $\bm{x}_{\mathrm{eprompt}}$ denote the set of reference captions and the evaluation prompt, respectively. Then $E^{(j)}$ outputs a score $\psi_{E^{(j)}}(\bm{x}_{e}^{(i,k)}) \in [0,1]$. Subsequently, we construct $\mathrm{\Phi} \in\mathbb{R}^{M\times M}$, whose $(i,j)$-th entry is as follows: 
\begin{equation*}
    \mathrm{\Phi}_{ij} = \frac{1}{N} \sum_{k=1}^{N} \psi_{E^{(j)}}(\bm{x}_{e}^{(i,k)}).
\end{equation*}
\vspace{-2mm}

\para{Step3. Row-and-column standardization} To compare the preferences across all Evaluator–Generator pairs, we obtain $\tilde{\mathrm{\Phi}} =\lbrack \tilde{\phi}_{ij} \rbrack \in \mathbb{R}^{M \times M}$, by sequentially standardizing $\mathrm{\Phi}$ in the column-wise and row-wise directions. To investigate model-specific preference bias, a direct comparison of the entries in $\mathrm{\Phi}$ is insufficient because the mean and standard deviation of the entries vary across both columns (i.e., Evaluators) and rows (i.e., Generators).

Therefore, we convert $\mathrm{\Phi}$ into $\tilde{\mathrm{\Phi}}$ through the following two steps: First, $\mathrm{\Phi}$ is standardized column-wise so that all columns have the same mean and standard deviation. Next, the matrix is standardized row-wise to isolate the preference relationships among the MLLMs, which yields $\tilde{\mathrm{\Phi}}$.

Each element $\tilde{\phi}_{ij}$ of $\tilde{\mathrm{\Phi}}$ corresponds to the $j$-th model’s evaluation of $\{ \hat{\bm{y}}_{g}^{(i,k)} \}_{k=1}^{N}$, following Evaluator- and Generator-wise standardization. For convenience, we use the notation $\tilde{\mathrm{\Phi}}_{j}(i)$ to refer to $\tilde{\phi}_{ij}$. In particular, $\tilde{\mathrm{\Phi}}_{i}(i)$ indicates the degree of self-preference, which we name the \textit{philautia score} ($\mathrm{\Phi}$-score). As described in Sections \ref{sec:intro} and \ref{sec:related}, existing methods fail to disentangle model performance from preference bias. Therefore, we use $\tilde{\mathrm{\Phi}}_{j}(i)$ for the direct and quantitative comparison of model-specific preferences across all possible Evaluator–Generator pairs.
\begin{algorithm}[t]
\caption{Philautia-Eval}
\label{alg:xprefeval}
\begin{algorithmic}[1]
\State \textbf{Input: }  Generators $\mathcal{G} = \{G^{(i)}\}$, Evaluators $\mathcal{E} = \{E^{(j)}\}$, Images $\{\bm{x}_{\text{img}}^{(k)}\}$, References $\{\bm{X}_{\text{ref}}^{(k)}\}$, Generation prompt $\bm{x}_{\text{gprompt}}$,  Evaluation prompt $\bm{x}_{\text{eprompt}}$
\State $\hat{\bm{y}}_g^{(i,k)} \gets \textsc{Generate}(G^{(i)}, \bm{x}_{\text{img}}^{(k)}, \bm{x}_{\text{gprompt}}),\forall i,k$ 

\For{each $(G^{(i)}, E^{(j)}) \in \mathcal{G} \times \mathcal{E}$} \Comment{\textcolor{blue}{Caption evaluation}}
    \State $\mathrm{\Phi}[i,j] \gets \textsc{Average}_k (\textsc{Eval}(\bm{x}_{\text{img}}^{(k)}, \bm{X}_{\text{ref}}^{(k)}, \hat{\bm{y}}_g^{(i,k)}, \bm{x}_{\text{eprompt}}))$
\EndFor
\State $\tilde{\mathrm{\Phi} }[:,j] \gets \textsc{Standardize}(\mathrm{\Phi} [:,j]),\forall j$ \Comment{\textcolor{blue}{Evaluator-wise}}
\State $\tilde{\mathrm{\Phi}}[i,:] \gets \textsc{Standardize}(\tilde{\mathrm{\Phi} }[i,:]),\forall i$ \Comment{\textcolor{blue}{Generator-wise}}
\State \Return $\tilde{\mathrm{\Phi} }$
\vspace{2mm}
\end{algorithmic}
\end{algorithm}

\vspace{-3mm}
\subsection{\textsc{Pomms}}
As shown in \cref{sec:exp}, most MLLMs exhibit model-specific preference bias. To construct a fair Evaluator from such MLLMs, we consider aggregating the evaluations from multiple Evaluators rather than relying on a single Evaluator. Specifically, we adopt a simple ensemble approach termed \textsc{Pomms}.

We select the component models (ensemble members) and optimize their coefficients as follows: The sequential feature selection algorithm (SFS) is used for the model selection. We should not apply complex deep neural networks to optimize their coefficients because of the low dimensionality of the input space (up to $M$ evaluations). Therefore, we employ a lightweight meta-learner (e.g., Elastic Net). A lower philautia score for \textsc{Pomms} than for individual Evaluators indicates a successful reduction in model-specific preference bias.

\vspace{-3mm}
\section{SelfEval-Cap Dataset}

To investigate model-specific preference bias in MLLM-as-a-Judge, we constructed the SelfEval-Cap dataset.
Addressing the research questions stated in \cref{sec:problemset}, we require a dataset that includes both generated captions and evaluation scores from a sufficiently large and diverse set of MLLMs. However, existing datasets (e.g.,~\cite{mllm-as-a-judge}) do not cover a sufficient number of MLLMs to satisfy this requirement. Consequently, it is difficult to investigate shared preference tendencies across MLLMs (e.g., patterns among MLLMs that share the same backbone LLM). Moreover, these datasets are not suitable for investigating the mitigation effects of ensembles proposed in \cref{sec:method}, because this requires a sufficient set of candidate Evaluators.

To address these limitations, we constructed the SelfEval-Cap dataset, which contains 54,000 captions generated by 12 MLLMs, and 1,296,000 evaluation scores given by the MLLMs. The SelfEval-Cap dataset covers a significantly larger and more diverse set of MLLMs than previous datasets (e.g., \cite{mllm-as-a-judge}). This diverse collection includes multiple variants within the LLaVA family (e.g., LLaVA-OneVision-7B~\cite{llavaov}) and Qwen-based MLLMs (e.g., Qwen2.5-VL-7B~\cite{qwenvl25}, Molmo-7B-D-0924~\cite{molmo}), as well as proprietary models (e.g., GPT-4o \cite{gpt4o}).

For data curation, we used images and human-provided captions in the validation set of the nocaps dataset~\cite{nocaps}, which is widely used in image captioning~\cite{polos, qwen-vl}. 
The nocaps dataset does not contain captions generated by MLLMs.
Therefore, we generated 54,000 captions for the images using 12 different MLLMs as Generators. Subsequently, we employed the MLLMs as Evaluators to evaluate the generated captions. These evaluations were performed in both reference-based and reference-free settings, resulting in a total of 1,296,000 scores.

\vspace{-1mm}
\para{Models}
We used the following MLLMs as both Generators and Evaluators: GPT-4o \cite{gpt4o}, Gemini 2.5 Pro \cite{gemini25pro}, Gemma-3-4B-IT \cite{gemma3}, DeepSeek-VL2 \cite{deepseekvl2}, LLaVA-1.5-13B \cite{llava-1.5}, LLaVA-NeXT-Vicuna-7B \cite{llavanext}, LLaVA-OneVision-7B \cite{llavaov}, Phi-3.5-Vision \cite{phi35vision}, Qwen2.5-VL-7B \cite{qwenvl25}, InternVL2.5-8B \cite{internvl25}, Eagle2-9B \cite{eagle2}, and Molmo-7B-D-0924 \cite{molmo}. For each model, the parameter sizes, backbone models (vision encoders and language models), hyperparameters, and further details are provided in Appendix \ref{sec:appendix_setup}. We selected these models because they are representative MLLMs \cite{mllm_survey1, mllm_survey2, mllm_survey3, vlm_survey}. Moreover, we considered model diversity in terms of  the organizations providing the MLLMs and model families. For inference by open MLLMs, we used an NVIDIA H200 SXM GPU (141 GB VRAM) for Eagle2-9B, Molmo-7B-D, and DeepSeek-VL2, and a GeForce RTX 4090 GPU (24 GB VRAM) for the other MLLMs.

\para{Prompts}
We employed the following prompts for the Generators and Evaluators. The Generators adopted the prompt used in LLaVA-OneVision: ``Provide a one-sentence caption for the provided image.'' In contrast, the Evaluators employed the prompt from G-VEval \cite{tong2024gveval}, because this is a representative MLLM-as-a-Judge. Although this prompt requests a score on a $0$-$100$ scale, we linearly normalized all evaluation scores to $[0,1]$. The full prompts can be found in Appendix~\ref{sec:appendix_prompt}.

\para{Statistics} 
The SelfEval-Cap dataset contains 4,500 images and 54,000 captions generated by 12 models, each generating a single caption in English per image. Each of the captions was evaluated by 12 Evaluators, in both reference-based and reference-free settings, resulting in a total of 1,296,000 scores. Moreover, the dataset includes 45,000 human-written captions that are used as references.
These captions were curated from the nocaps dataset and have a vocabulary size of 11,404 words, a total word count of 505,721, and an average length of 11.24 words.

\section{Experimental Results}
\label{sec:exp}
\subsection{RQ1: To What Extent Does MLLM-as-a-Judge Exhibit Self-Preference Bias?}
\vspace{-2mm}
\begin{figure}[!t]
    \centering
    \includegraphics[width=\linewidth]{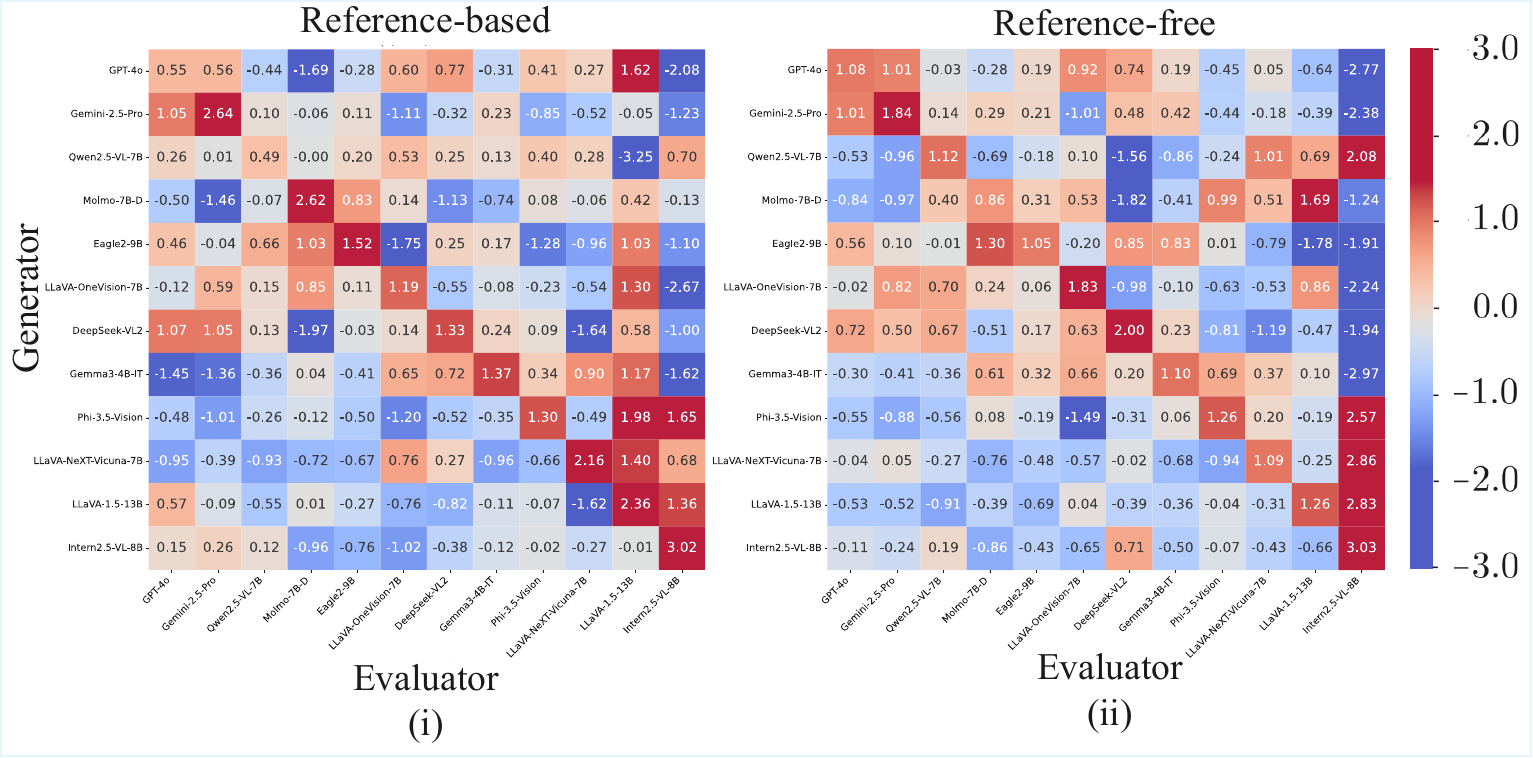}
    \caption{\textbf{Visualization of} $\tilde{\mathrm{\Phi}}$ \textbf{in the (i) reference-based and (ii) reference-free settings.} All philautia scores (diagonal items) were greater than zero, indicating the presence of self-preference bias within the MLLMs used in our experiments. }
    \label{fig:standardized_matrix}
    \vspace{-3mm}
\end{figure}

\para{Representative MLLMs tend to exhibit self-preference bias}
We investigated RQ1 by visualizing $\tilde{\mathrm{\Phi}}$ and comparing the philautia scores across Evaluators. \cref{fig:standardized_matrix}-(i) shows $\tilde{\mathrm{\Phi}}$ in the reference-based setting. All philautia scores were greater than zero, indicating the presence of self-preference bias in the MLLMs used in our experiments. InternVL2.5-8B exhibited the highest philautia score of 3.02. Furthermore, it favored LLaVA-1.5-13B and Phi-3.5-Vision while disfavoring LLaVA-OneVision-7B and GPT-4o. These findings suggest that InternVL2.5-8B has a relatively strong self-preference bias and that it tends to give extreme scores to other models. A similar tendency was observed in DeepSeek-VL2; specifically, it favored itself and GPT-4o, while disfavored Molmo-7B-D.

The extent of the bias is further indicated by the magnitude of the deviation from the mean. In the case of Gemma3-4B-IT, the philautia score was 1.37, whereas the mean of $\{\tilde{\mathrm{\Phi}}_{\text{Gemma3}}(G^{(i)})\}_{i=1}^M$ was $-0.0478$. This philautia score was more than two standard deviations from the mean of  $\{\tilde{\mathrm{\Phi}}_{\text{Gemma3}}(G^{(i)})\}_{i=1}^M$. Such a substantial deviation indicates that Gemma3-4B-IT also has a relatively strong self-preference bias. Similarly, Eagle2-9B and Gemini-2.5-Pro demonstrated relatively strong self-preference bias, as supported by the significant deviations of their philautia scores from their respective means. Specifically, the philautia scores for these models were 1.52 and 2.64, respectively. These values were also more than two standard deviations from the mean of $\{\tilde{\mathrm{\Phi}}_{\text{Eagle2}}(G^{(i)})\}_{i=1}^M$ and $\{\tilde{\mathrm{\Phi}}_{\text{Gemini-2.5-Pro}}(G^{(i)})\}_{i=1}^M$. 

\cref{fig:standardized_matrix}-(ii) shows $\tilde{\mathrm{\Phi}}$ in the reference-free setting.
The subfigure shows that all philautia scores exceeded zero, indicating that the MLLMs exhibit self-preference bias in the reference-free setting as well. 
In particular, InternVL2.5-8B has the highest philautia score in both settings, with a difference of less than 0.01 between them. 

\para{References affect self-preference bias in Gemini 2.5 Pro} 
Compared with the reference-based setting (\cref{fig:standardized_matrix}-(i)), the philautia scores in the reference-free setting (\cref{fig:standardized_matrix}-(ii)) were higher for LLaVA-OneVision-7B (rising from 1.19 to 1.83), Qwen2.5-VL-7B (0.49 to 1.12), DeepSeek-VL2 (1.33 to 2.00), and GPT-4o (0.55 to 1.08). 
In contrast, the scores exhibit substantial reductions for Molmo-7B-D (dropping from 2.62 to 0.86), LLaVA-1.5-13B (2.36 to 1.26), LLaVA-NeXT-Vicuna-7B (2.16 to 1.09), and Gemini-2.5-Pro (2.64 to 1.84).
These results indicate that in most cases, the presence of references significantly affects self-preference bias. 
Therefore, if we were to use reference-based MLLM-as-a-Judge, the references would need to be carefully designed.

\para{GPT-4o has a relatively low self-preference bias}
As shown in \cref{fig:standardized_matrix}-(i), Qwen2.5-VL-7B and GPT-4o exhibited relatively low philautia scores, at 0.49 and 0.55, respectively.
\begin{figure}[!t]
    \centering
    \includegraphics[width=\linewidth]{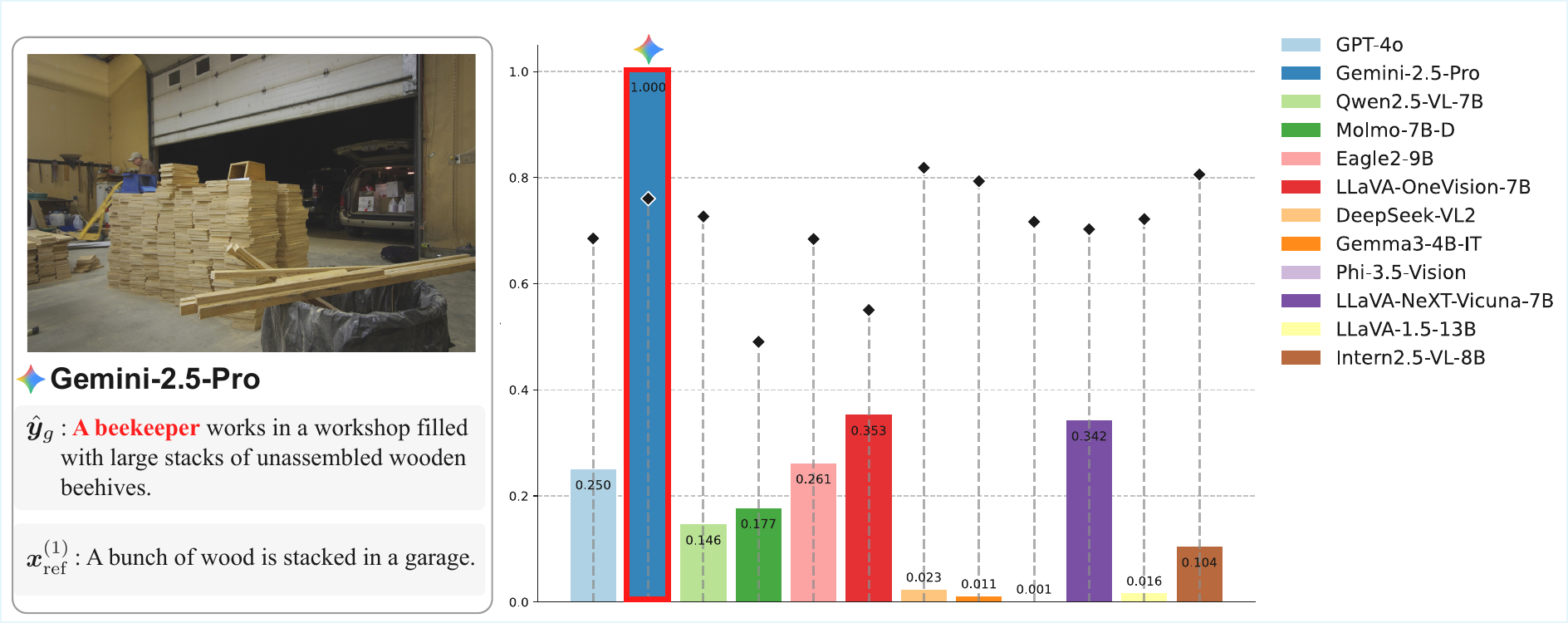}
    \caption{\textbf{Example of self-preference bias.} The bar chart shows the scores given to a caption generated by Gemini-2.5-Pro.  Gemini-2.5-Pro exceptionally gave high scores to its own generations compared with the other Evaluators. The symbol $\blacklozenge$ represents the mean value of the scores by each Evaluator. \textbf{\textcolor{red}{Red}} text within $\hat{\bm{y}}_{g}$ highlights hallucination.}
    \label{fig:qualitive-1}
    \vspace{-4mm}
\end{figure}
Although Qwen2.5-VL-7B appears to have the lowest philautia score, this value might be skewed by the evaluations by LLaVA-1.5.
Most $\tilde{\mathrm{\Phi}}_{E^{(j)}}(\text{Qwen2.5-VL})$ values were greater than zero, whereas LLaVA-1.5 disfavored Qwen2.5-VL, giving it a significantly low score 
($\tilde{\mathrm{\Phi}}_{\text{LLaVA1.5}}(\text{Qwen2.5-VL}) = -3.25$), which was the lowest value even within the entire $\tilde{\mathrm{\Phi}}$. 
This outlier increased the row-wise variance, causing the Generator-wise standardization to yield a low philautia score. 
Indeed, when we computed $\tilde{\mathrm{\Phi}}$ without LLaVA-1.5, the philautia score for Qwen2.5-VL-7B was 0.934.
In contrast, GPT-4o was the only model whose philautia score was less than one standard deviation from the mean of $\{\tilde{\mathrm{\Phi}}_{\text{GPT-4o}}(G^{(i)})\}_{i=1}^M$. This indicates that GPT-4o has a relatively low self-preference bias. A visualization of the raw score heatmaps and the related discussion are provided in Appendix~\ref{sec:appendix_raw_matrix}.

\para{Qualitative results}
\cref{fig:qualitive-1} shows a case in which MLLMs evaluated a caption generated by Gemini-2.5-Pro. This caption incorrectly described the wood as beehives and mentioned a beekeeper who is not present in the image. 
While most Evaluators $E^{(j)}$ gave scores lower than their respective mean values to this caption, Gemini-2.5-Pro exceptionally gave a significantly high score of 1.000 to its own generation.
This score was 0.239 points higher than the mean value of scores for itself (i.e., 0.761), indicating that Gemini-2.5-Pro favored its own generation despite the errors. 
Additional qualitative results are provided in Appendix \ref{sec:appendix_qualitative}. 

\vspace{-3mm}
\subsection{RQ2: To What Extent Does Cross-Model Preference Bias Appear in MLLM-as-a-Judge?}
\vspace{-1mm}
\begin{figure}[!t]
    \centering
    \includegraphics[width=0.9\linewidth]{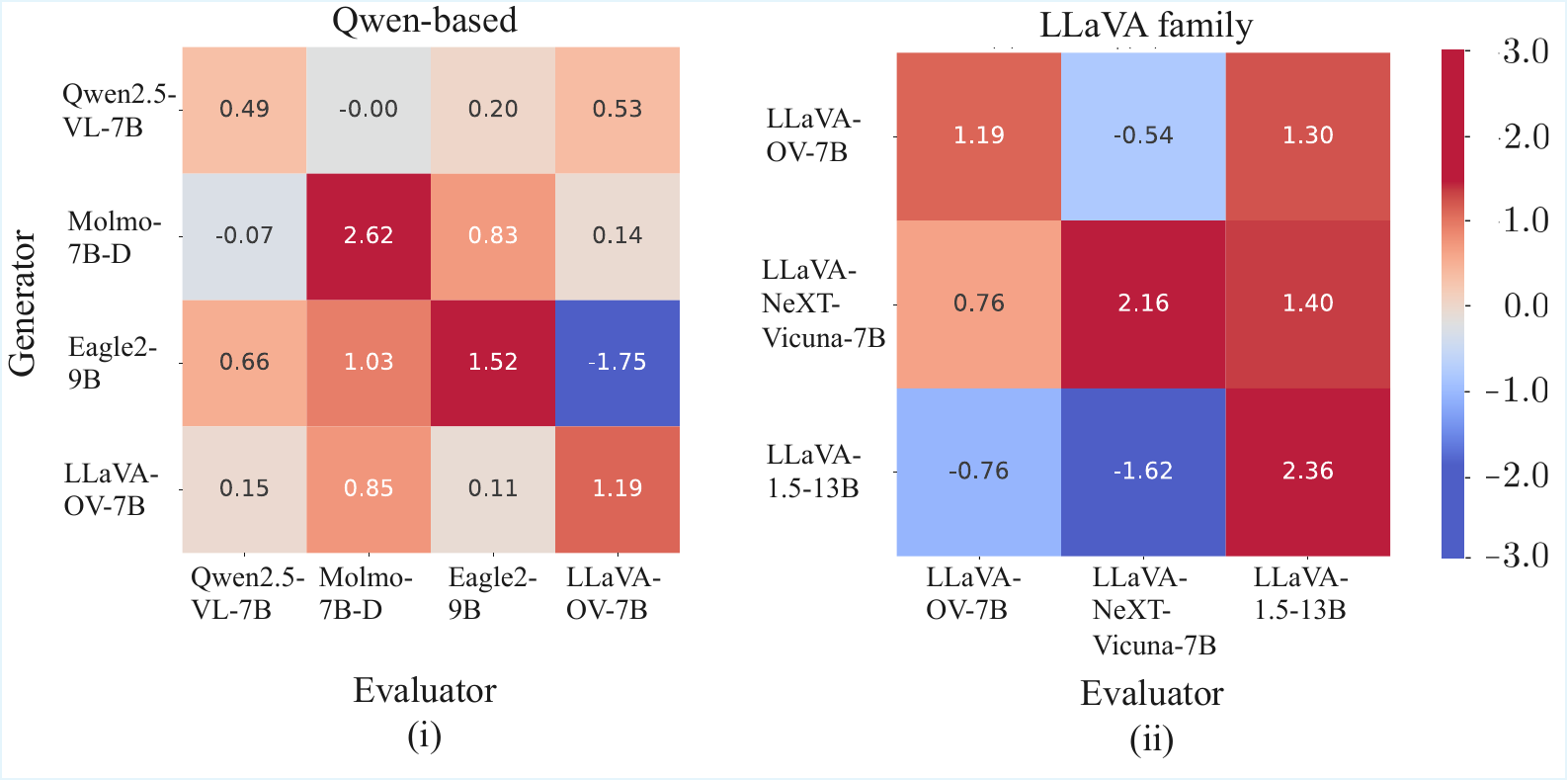}
    \caption{\textbf{Visualization of preference bias within model families.} (i) Submatrix for Qwen-based models: nine of the 12 off-diagonal entries were positive, suggesting a preference bias within the model family. (ii) Submatrix for LLaVA-family models: LLaVA-1.5-13B tends to favor its successor models (e.g., LLaVA-NeXT-Vicuna-7B, LLaVA-OneVision-7B).}
    \label{fig:curated_matrix}
    \vspace{-3mm}
\end{figure}

We investigated preference relationships within particular model families (e.g., Qwen-based MLLMs and LLaVA family).
\cref{fig:curated_matrix}-(i) shows the $4 \times 4$ submatrix of $\tilde{\mathrm{\Phi}}$ for the Qwen-based models, while \cref{fig:curated_matrix}-(ii) shows its $3 \times 3$ submatrix for the LLaVA-family models. 

\para{Qwen-based MLLMs tend to favor each other}
As shown in subfigure~(i), 9 out of 12 off-diagonal entries were greater than zero in the $4\times4$ submatrix for the Qwen-based models.
This matrix had the second-largest number of positive off-diagonal entries, among all $\binom{12}{4}$ possible $4\times4$ submatrices.
These findings suggest that Qwen-based models tend to favor each other.

\para{Within the LLaVA family, LLaVA-1.5 tends to favor its successor models}
\cref{fig:standardized_matrix}~(ii) shows that $\tilde{\mathrm{\Phi}}_{\mathrm{LLaVA\text{-}1.5}}(\mathrm{LLaVA\text{-}NeXT})$ and $\tilde{\mathrm{\Phi}}_{\mathrm{LLaVA\text{-}1.5}}(\mathrm{LLaVA\text{-}OV})$ were relatively high at 1.40 and 1.30, respectively, compared with the other off-diagonal entries. This is likely because LLaVA-NeXT-Vicuna-7B and LLaVA-OneVision-7B are successor models to LLaVA-1.5-13B, and they inherit substantial parts of the LLaVA-1.5 training pipeline. Specifically, LLaVA-NeXT-Vicuna-7B reuses the pretrained connector from LLaVA-1.5 and leverages LLaVA-1.5 instruction-tuning data during visual instruction tuning~\cite{llavanext}. 
Similarly, LLaVA-OneVision builds on LLaVA-NeXT-derived resources, including large-scale recaptioned detailed-description data generated by LLaVA-NeXT-34B~\cite{llavaov}.
These results suggest that LLaVA-1.5-13B tends to favor its successor models (e.g., LLaVA-NeXT-Vicuna-7B, LLaVA-OneVision-7B).

\vspace{-4mm}
\subsection{RQ3. Can an Ensemble of Evaluators Mitigate the Influence of Model-Specific Preference Bias While Maintaining Alignment with Human Judgments?}
\label{sec:rq3}
\begin{table}[!t]
\renewcommand*{\arraystretch}{1.25}
\setlength{\tabcolsep}{8pt}
\newcolumntype{M}[1]{>{\centering\arraybackslash}m{#1}}
\centering
\caption{\textbf{Quantitative comparison between \textsc{Pomms} and the baselines.} For \textsc{Pomms}, each row sequentially adds a model to the ensemble. The philautia scores for G-VEval are obtained from the $\tilde{\mathrm{\Phi}}$ matrix in \cref{fig:standardized_matrix}~(ii); therefore, the values are not identical to those calculated from $\tilde{\mathrm{\Phi}}$ containing \textsc{Pomms}. The details are provided in Appendix~\ref{sec:appendix_results}. Bold font indicates the best results, and underlined font indicates the second-best results. ``$\mathrm{\Phi}$-score'' represents the philautia score. }
\scalebox{0.9}{
\begin{tabular}{llccccc}
\toprule
\multicolumn{2}{l}{\textbf{Metrics}} &
\multicolumn{2}{c}{\textbf{Nebula}} & \multicolumn{2}{c}{\textbf{Flickr8k-Ex}} & \multicolumn{1}{c}{\textbf{SelfEval-Cap}} \\
\cmidrule(lr){3-4} \cmidrule(lr){5-6} \cmidrule(lr){7-7}
\multicolumn{2}{l}{} & 
$\tau_b$ $\uparrow$ & $\tau_c$ $\uparrow$ & $\tau_b$ $\uparrow$ & $\tau_c$ $\uparrow$ & $\mathrm{\Phi}$-score \\
\midrule

\multirow{3}{*}{\rotatebox[origin=c]{90}{\footnotesize \textbf{G-VEval}}} &
GPT-4o & 56.1 & 53.2 & \underline{61.5} & \underline{59.7} & 1.08 \\
&  
Qwen2.5-VL-7B & 55.3 & 52.4 & 54.6 & 54.0 & 1.12 \\
&  
InternVL2.5-8B & 54.1 & 51.3 & 54.6 & 52.9 & 3.03 \\
\midrule

\multirow{6}{*}{\rotatebox[origin=c]{90}{\footnotesize \textbf{\textsc{Pomms}}}}   &(i)
\begin{tabular}[c]{@{}l@{}}GPT-4o and InternVL2.5-8B\end{tabular}
& 56.4 & 53.5 & \underline{61.5} & \underline{59.7} & 1.31 \\
  &(ii)
+ Eagle2-9B & \underline{56.6} & 53.6 & \textbf{62.7} & \textbf{60.8} & 0.45 \\
 &(iii) 
+ LLaVA-OneVision-7B & \underline{56.6} & \underline{53.7} & 60.6 & 58.8 & \underline{-0.19} \\
 &(iv) 
+ DeepSeek-VL2 & 56.4 & 53.5 & 59.0 & 57.3 & \textbf{0.15} \\
 &(v) 
+ Qwen2.5-VL-7B & \textbf{57.0} & \textbf{54.1} & 59.6 & 57.8 & 0.52 \\
 &(vi) 
+ Phi-3.5-Vision & \textbf{57.0} & \textbf{54.1} & 59.6 & 57.8 & 0.42 \\
\bottomrule
\end{tabular}
}
\label{tab:quantitative_full_main}
\end{table}

\para{Setup}
To validate \textsc{Pomms}, we used Flickr8k-Expert~\cite{flickr} and Nebula~\cite{deneb}, which are standard benchmarks for image captioning metrics. Following prior studies \cite{pac-s, pac-spp, tong2024gveval, lee2024fleur, polos, vela}, we adopted Kendall’s $\tau_b$ and $\tau_c$, as evaluation metrics. For the baselines, we adopted G-VEval \cite{tong2024gveval} based on GPT-4o and its variants equipped with Qwen2.5-VL-7B and InternVL2.5-8B backbones.  We selected G-VEval because it is a representative MLLM-as-a-Judge for image captioning. 

The statistics of the Flickr8k-Expert and Nebula datasets are as follows: The Flickr8k-Expert dataset comprises 1,000 images, 16,992 human judgments collected from 21 annotators. The Nebula dataset contains 32,978 samples. Each sample includes an image, a candidate caption, a set of reference captions, and a human judgment. These human judgments were collected from 805 annotators, and the benchmark contained a total of 32,978 unique images. Details of these datasets are provided in Appendix \ref{sec:appendix_setup}. We optimized the \textsc{Pomms}’ ensemble weights using the training set of Nebula~\cite{deneb}. We followed the data split defined in the Nebula dataset, where the training, validation, and test sets contained 26,382, 3,298, and 3,298 samples, respectively. We used the training set for training the ensemble and the validation set for model selection for the ensemble. The test set was used for validating the models’ performance.

\para{\textsc{Pomms} mitigates preference bias while maintaining performance}
Table \ref{tab:quantitative_full_main} presents a quantitative comparison between \textsc{Pomms} and the baselines on the Flickr8k-Expert, Nebula, and SelfEval-Cap benchmarks. As shown in the table, \textsc{Pomms} (iv) achieved a significantly lower philautia score of $0.15$ compared to the baselines on SelfEval-Cap. \textsc{Pomms} (iv) showed a $\tau_b$ score of 56.4 on the Nebula dataset, outperforming G-VEval (GPT-4o) by 0.3 points. Moreover, it performed comparably to the baselines on the Flickr8k-Expert dataset, with a $\tau_b$ score of 59.0. These results suggest that \textsc{Pomms} (iv) effectively mitigated the influence of self-preference bias while maintaining performance.
\begin{table}[!t]
\centering
\begingroup
\setlength{\aboverulesep}{0pt}
\setlength{\belowrulesep}{0pt}
\renewcommand*{\arraystretch}{1.25}
\setlength{\tabcolsep}{3pt}
\caption{
\textbf{Quantitative comparison between} $\tilde{\mathrm{\Phi}}_{E^{(i)}}(G^{(i)})$ \textbf{and} $\tilde{\mathrm{\Phi}}_{\text{\textsc{Pomms}}}(G^{(i)})$. Bold font indicates the best results. The philautia scores are computed in the same manner as in Table~\ref{tab:quantitative_full_main}. The full model names for the abbreviations are as follows: Gemini: Gemini-2.5-Pro; Qwen: Qwen2.5-VL-7B; Molmo: Molmo-7B-D; Eagle: Eagle2-9B; LV-OV: LLaVA-OneVision-7B; DeepSeek: DeepSeek-VL2; Gemma: Gemma3-4B-IT; Phi: Phi-3.5-Vision; LV-NV: LLaVA-NeXT-Vicuna-7B; LV: LLaVA-1.5-13B; Intern: InternVL2.5-8B.
}
\vspace{-3mm}

\scalebox{0.83}{
\begin{tabular}{lcccccccccccc}
\toprule
\multirow{2}{*}{\textbf{Evaluator}} &
\multicolumn{12}{c}{\textbf{Generator}} \\
\cmidrule(lr){2-13}
& \scriptsize GPT-4o
& \scriptsize Gemini
& \scriptsize Qwen
& \scriptsize Molmo
& \scriptsize Eagle
& \scriptsize LV-OV
& \scriptsize DeepSeek
& \scriptsize Gemma
& \scriptsize Phi
& \scriptsize LV-NV
& \scriptsize LV
& \scriptsize Intern \\
\midrule
Self & 1.08 & 1.84 & 1.12 & 0.86 & 1.05 & 1.83 & 2.00 & 1.10 & 1.26 & 1.09 & 1.26 & 3.03 \\
\textsc{Pomms}   & \textbf{0.23} & \textbf{0.18} & \textbf{0.67} & \textbf{-0.26} & \textbf{0.01} & \textbf{-0.90} & \textbf{0.67} & \textbf{-0.33} & \textbf{0.11} & \textbf{0.14} & \textbf{-0.18} & \textbf{0.34} \\
\bottomrule
\end{tabular}}
\label{tab:phi_mometrics}
\vspace{-3mm}
\endgroup
\end{table}

A similar trend was observed in the sensitivity analysis presented in Table \ref{tab:quantitative_full_main}, when we increased the number of models in \textsc{Pomms}. \textsc{Pomms} (ii)-(vi) demonstrated lower philautia scores and comparable performance on $\tau_b$ and $\tau_c$, compared with the baselines. These results indicate that the model selection based on SFS is simple but sufficiently reasonable. 

Table \ref{tab:phi_mometrics} presents the degree of model preferences of \textsc{Pomms}, against other MLLMs ($\tilde{\mathrm{\Phi}}_{\text{\textsc{Pomms}}}(G^{(i)})$). We compared this against the philautia scores of the MLLMs. This instance of \textsc{Pomms} refers to \textsc{Pomms} (iv). We adopted this combination of MLLMs because, as shown in Table \ref{tab:quantitative_full_main}, it represents the mixture with the smallest absolute philautia score. Table \ref{tab:phi_mometrics} demonstrates the scores given by \textsc{Pomms} were consistently and significantly lower than the corresponding philautia scores of the component models (i.e., GPT-4o, InternVL2.5-8B, Eagle2-9B, LLaVA-OneVision-7B, and DeepSeek-VL2). Moreover, this reduction in score magnitude was also observed for models not included in the \textsc{Pomms} components.

Even the most extreme score given by \textsc{Pomms} ($\tilde{\mathrm{\Phi}}_{\text{\textsc{Pomms}}}(\text{LLaVA-OV})=-0.90$) is smaller in magnitude than the smallest value among the most extreme scores from each Evaluator ($\tilde{\mathrm{\Phi}}_{\text{Eagle2}}(\text{Eagle2})=1.05$). This result indicates that \textsc{Pomms} is less likely to give extreme scores than the individual Evaluators. Taken together with Table \ref{tab:quantitative_full_main}, these results indicate that \textsc{Pomms} mitigated the influence of model-specific preference bias.

\vspace{-4mm}
\subsection{Discussion}
We observed strong mutual preference biases between specific model pairs (e.g., GPT-4o~\cite{gpt4o} and DeepSeek-VL2~\cite{deepseekvl2}).
This hints at a potential homogenization of alignment standards or circular data dependencies in the current (M)LLM landscape. 
However, identifying the definitive causes of such preferences is inherently difficult. 
Because the detailed architectures and training data of proprietary models are undisclosed, we cannot determine whether these correlations arise from shared training corpora or similar architectural designs. 
Consequently, disentangling these complex inter-model dynamics remains a challenge for future research.
Furthermore, as LLM-as-a-Judge becomes standard practice, establishing an acceptable threshold for such bias requires community consensus.

\para{Limitations} We demonstrated model-specific preference bias in image captioning. However, our findings do not directly demonstrate model-specific preference bias in other vision-language tasks (e.g., VQA). 
In future work, we plan to investigate model-specific bias in other vision-language tasks (e.g., VQA) using Philautia-Eval. 
Furthermore, \textsc{Pomms} is computationally demanding at inference time, because it involves multiple component models.
Therefore, we will attempt to make \textsc{Pomms} more computationally efficient. 
Specifically, as the current ensemble approach requires multiple models to be queried, we plan to distill the aggregated preferences into a single, lightweight student model.

\vspace{-4mm}
\section{Conclusion}
In this study, we investigated the extent of self-preference bias in MLLM-as-a-Judge for image captioning. Furthermore, we examined whether an aggregation of multiple MLLM-as-a-Judge methods can mitigate the influence of model-specific preference bias.

Our key findings are as follows: 
\begin{itemize}
    \item  [$\bullet$]Representative MLLMs tend to exhibit self-preference bias in both reference-based and reference-free settings for image captioning.
    \item  [$\bullet$]Experimental results indicate mutual preference bias within particular model families, presumably driven by reused connectors and overlapping instruction-tuning resources.
    \item  [$\bullet$]We introduce a simple ensemble of MLLMs, \textsc{Pomms}, which effectively mitigates the influence of model-specific preference bias, while maintaining performance.
\end{itemize}

\para{Acknowledgements}
This work was partially supported by JSPS KAKENHI Grant Number 23K28168, JST Moonshot, and JSPS Fellows Grant Number JP25KJ2069.

\renewcommand{\thesection}{\Alph{section}}
\renewcommand{\thetable}{\Alph{table}}
\renewcommand{\thefigure}{\Alph{figure}}

\bibliographystyle{splncs04}
\bibliography{main}

\title{
\hspace{-1mm}MLLM-as-a-Judge Exhibits Model Preference Bias\hspace{-1mm}
}

\author{
Shuitsu Koyama\thanks{Equal contribution}
\and
Yuiga Wada\samethanks \and
Daichi Yashima \and
Komei Sugiura
}

\authorrunning{S.~Koyama et al.}

\institute{
Keio University, Japan
\\\email{\{koyamashu3, yuiga, ydaichi1207, komei.sugiura\}@keio.jp}
}

\maketitle
\section{Details of Experimental Setup}
\label{sec:appendix_setup}

\subsection{Generators and Evaluators}
We used the following MLLMs as both Generators and Evaluators:
GPT-4o ~\cite{gpt4o}, Gemini 2.5 Pro ~\cite{gemini25pro}, Gemma-3-4B-IT ~\cite{gemma3}, DeepSeek-VL2 ~\cite{deepseekvl2}, LLaVA-1.5-13B ~\cite{llava-1.5}, LLaVA-NeXT-Vicuna-7B ~\cite{llavanext}, LLaVA-OneVision-7B ~\cite{llavaov}, Phi-3.5-Vision ~\cite{phi35vision}, Qwen2.5-VL-7B ~\cite{qwenvl25}, InternVL2.5-8B ~\cite{internvl25}, Eagle2-9B ~\cite{eagle2}, and Molmo-7B-D-0924 ~\cite{molmo}. Table \ref{tab:model-lists} presents the parameter sizes and backbone models (vision encoders and language models) for each model. We set the MLLM hyperparameters as follows: The hyperparameters for the Evaluators were set to $\text{temperature}=1$ and $\text{top-}p=1$, following G-VEval~\cite{tong2024gveval}.

\begin{table}[H]
    \normalsize
    \newcommand*{\bhline}[1]{\noalign{\hrule height #1}}
    \caption{\textbf{Details of the MLLMs used in our experiments.} ``\#Param''  indicates the number of parameters.}
    \vspace{2mm}
    \centering
    \scalebox{0.85}{
    \begin{tabular}{llll}
    \toprule
    \textbf{Model} & \textbf{\#Param} & \textbf{Vision Encoder} & \textbf{LLM backbone} \\ 
    \midrule
    GPT-4o~\cite{gpt4o} & - & - & - \\
    Gemini 2.5 Pro~\cite{gemini25pro} & - & - & - \\
    Qwen2.5-VL-7B~\cite{qwenvl25} & 8.29B & ViT & Qwen2.5-7B \\
    Molmo-7B-D-0924~\cite{molmo} & 8.02B & CLIP ViT-L/14 & Qwen2-7B \\
    Eagle2-9B~\cite{eagle2} & 8.93B & Siglip-400M+ConvNext & Qwen2.5-7B-Instruct\\
    LLaVA-OneVision-7B~\cite{llavaov} & 7.06B & SigLIP-SO400M-384 & Qwen2-7B \\
    DeepSeek-VL2~\cite{deepseekvl2} & 27.5B& SigLIP-SO400M-384/SAMB & DeepSeekMoE-27B \\
    Gemma-3-4B-IT~\cite{gemma3} & 4.3B & SigLIP-400M & Gemma 3 \\
    Phi-3.5-Vision~\cite{phi35vision} & 4.2B & CLIP ViT-L/14 & Phi-3-Mini \\
    LLaVA-NeXT-Vicuna-7B~\cite{llavanext} & 7.06B & CLIP ViT-L/14 & Vicuna-7B-v1.5 \\
    LLaVA-1.5-13B~\cite{llava-1.5} & 13.4B & CLIP ViT-L/14 & Vicuna-13B \\
    InternVL2.5-8B~\cite{internvl25} & 8.1B & InternViT-300M-448px & InternLM2.5-7B-Chat \\
    \bottomrule
    \end{tabular}
    }
    \label{tab:model-lists}
\end{table}

\subsection{Datasets}
\para{Flickr8k-Expert} The Flickr8k-Expert benchmark~\cite{flickr} comprises 1,000 images and 16,992 human judgments collected from 21 annotators. The total number of references is 5,000 with a vocabulary size of 3,241, total word count of 59,011, and average sentence length of 11.80 words. Furthermore, the total number of candidate captions is 5,664, with a vocabulary size of 1,518, total word count of 67,489, and average sentence length of 11.92. 

\para{Nebula} The Nebula benchmark~\cite{deneb} includes 183,472 reference captions, featuring a vocabulary size of 32,870 and a total word count of 1,945,956, with an average sentence length of 10.61 words. Furthermore, it contains 32,978 candidate captions, which exhibit a vocabulary size of 3,695, a total word count of 288,922, and an average sentence length of 8.76 words. 
In both benchmarks, all captions were written in English.

\section{Additional Experimental Results}
\label{sec:appendix_results}

\subsection{Raw Score Heatmap}
\begin{figure}[!t]
    \centering
    \includegraphics[width=\linewidth]{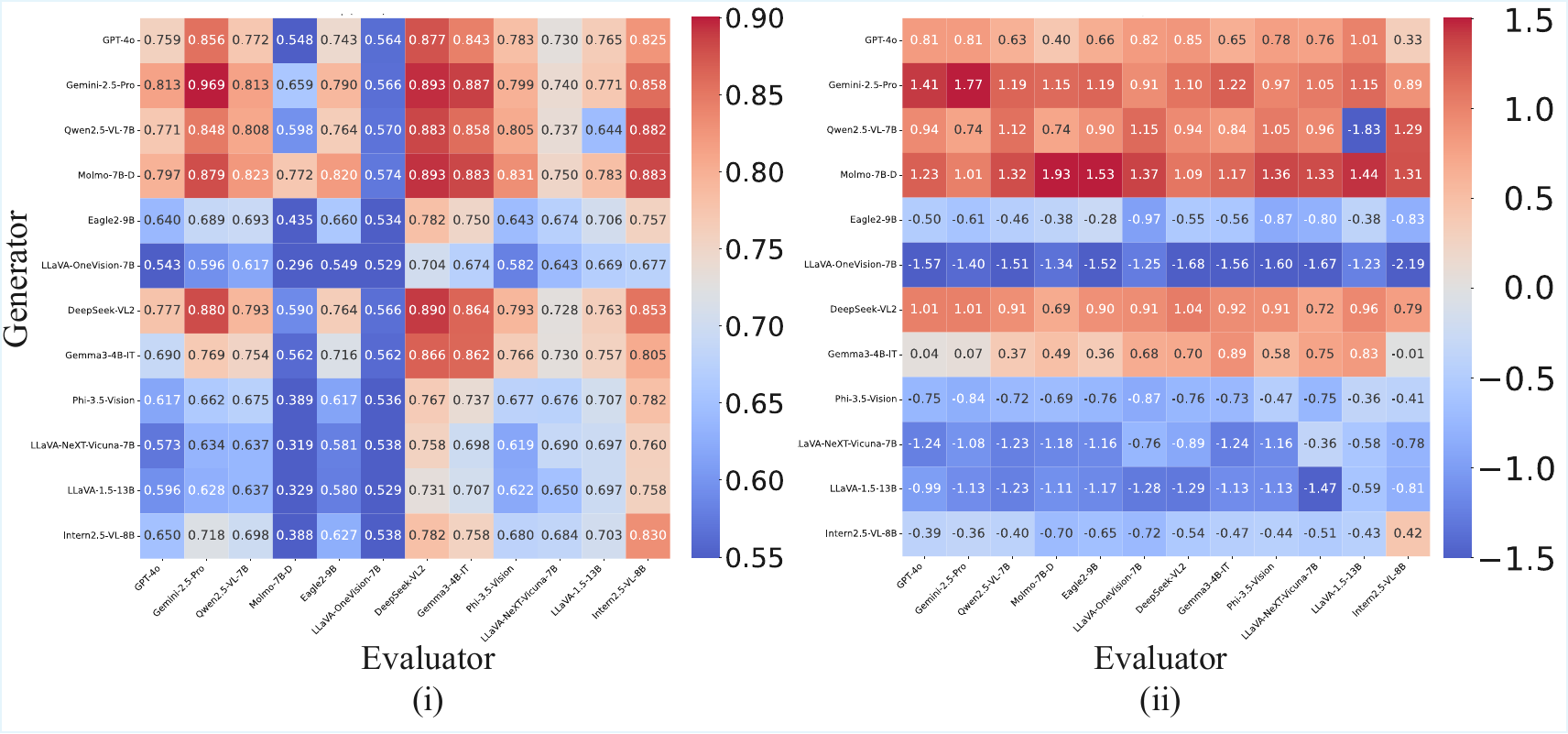}
    \caption{
     (i) \textbf{Visualization of} $\mathrm{\Phi}$ \textbf{in the reference-based setting.} The vertical and horizontal axes represent the Generators and Evaluators, respectively. High scores are shown in red and low scores in blue. (ii) \textbf{Visualization of} $\mathrm{\Phi}$ \textbf{after Evaluator-wise standardization.} Because the means and standard deviations differ across Evaluators, the matrix entries should be standardized in the columnwise direction. Similarly, they should be standardized in the row-wise direction.
     }
    \label{fig:raw_matrix}
    \vspace{-3mm}
\end{figure}

\label{sec:appendix_raw_matrix}

\cref{fig:raw_matrix}-(i) shows $\mathrm{\Phi}$ in the reference-based setting, while \cref{fig:raw_matrix}-(ii) shows $\mathrm{\Phi}$ standardized in the columnwise direction. In the subfigures, the vertical and horizontal axes represent the Generators and Evaluators, respectively. \cref{fig:raw_matrix}-(i) illustrates that both the mean and the standard deviation of $\mathrm{\Phi}_{E^{(j)}}(G^{(i)})$  vary substantially across Evaluators. For instance, the means of $\{\mathrm{\Phi}_{\text{DeepSeek-VL2}}(G^{(i)})\}_{i=1}^{M}$ and $\{\mathrm{\Phi}_{\text{Molmo}}(G^{(i)})\}_{i=1}^{M}$ were 0.819 and 0.490. Therefore, $\mathrm{\Phi}_{E^{(j)}}(G^{(i)})$ should be standardized in the columnwise direction. Furthermore, \cref{fig:raw_matrix}-(ii) shows that the row-wise means vary substantially because some MLLMs generate better captions and therefore have higher means than others. Consequently, the entries should be further standardized in the row direction to visualize the preference relationships among the MLLMs. 

\subsection{Qualitative Results}
\begin{figure}[!t]
    \centering
    \includegraphics[width=\linewidth]{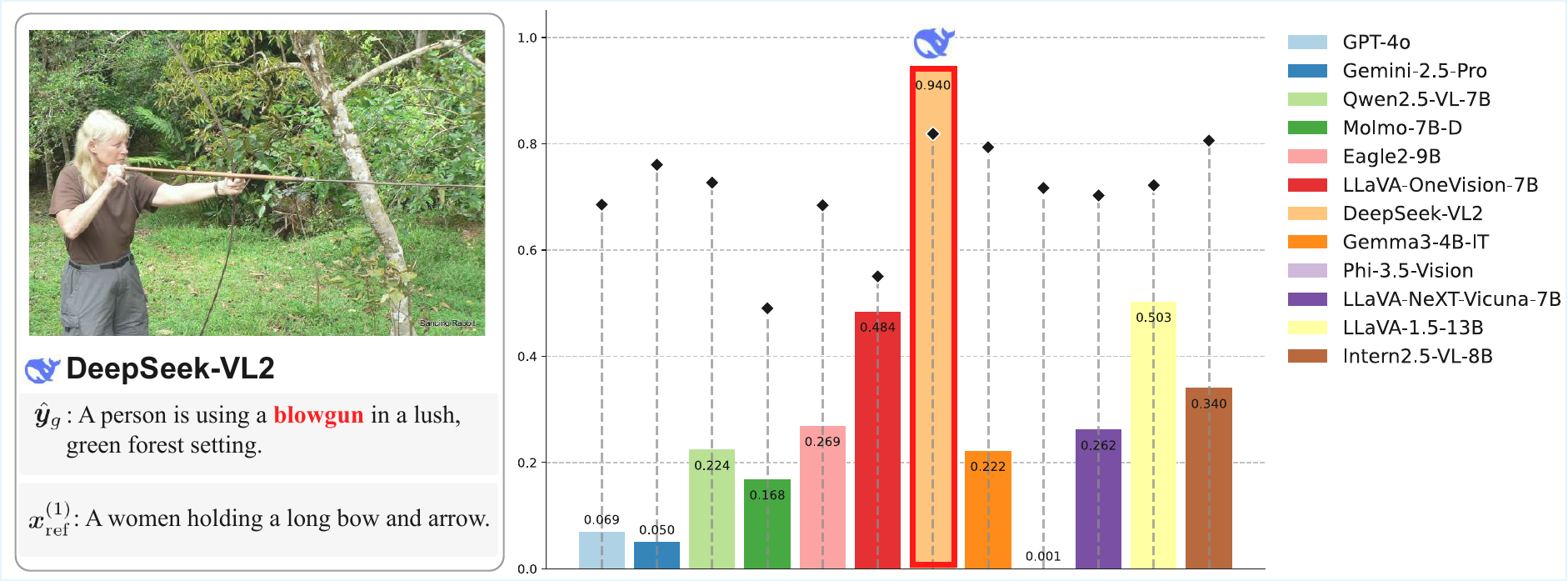}
    \caption{\textbf{Additional example of self-preference bias.} DeepSeek-VL2 exceptionally gave high scores to its own generations compared with the other Evaluators. The symbol $\blacklozenge$ represents the mean value of scores for each Evaluator. Red text within $\hat{\bm{y}}_{g}$ highlights hallucination.}
    \label{fig:qualitive-2}
    \vspace{-3mm}
\end{figure}

\begin{figure}[!t]
    \centering
    \includegraphics[width=\linewidth]{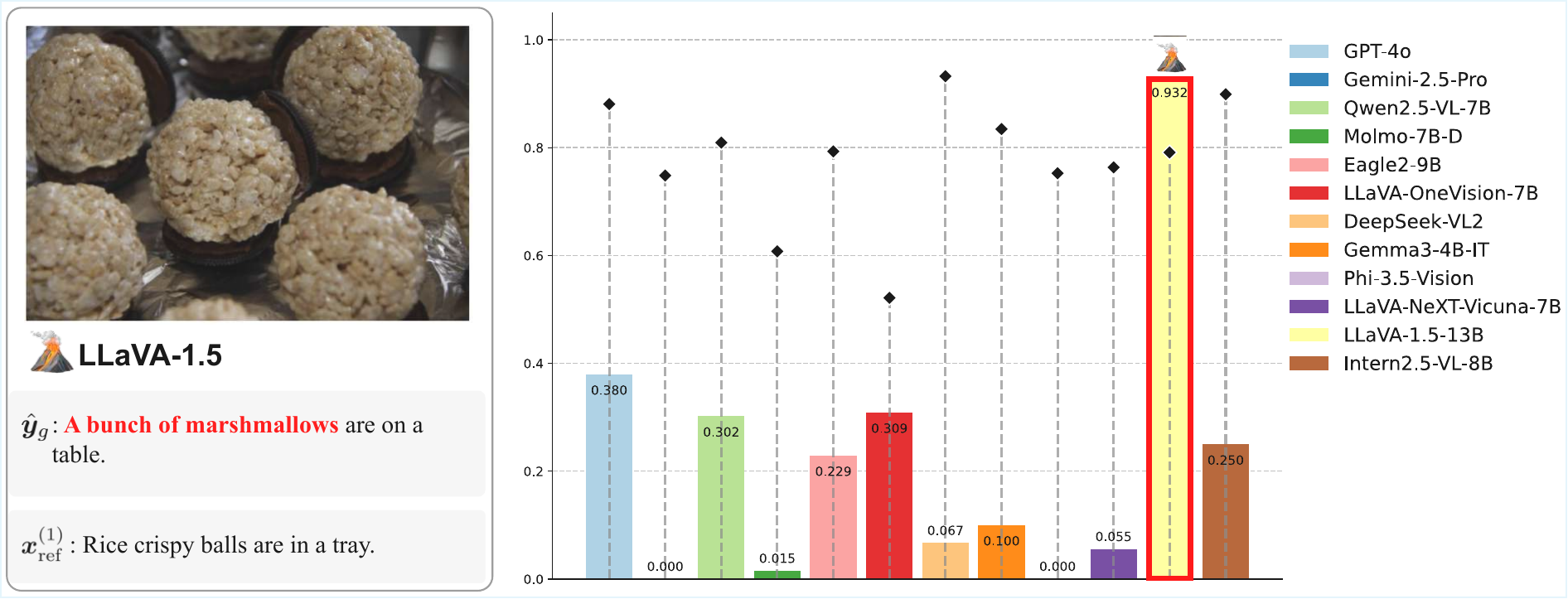}
    \caption{\textbf{Example of self-preference bias.} LLaVA-1.5 exceptionally gave high scores to its own generations compared with the other Evaluators. 
    }
    \label{fig:qualitive-3}
\end{figure}

\cref{fig:qualitive-2} shows an example of self-preference bias in DeepSeek-VL2. 
The figure shows the MLLM evaluations of a caption generated by DeepSeek-VL2 that contained a critical error: the model incorrectly identified a bow as a blowgun. Indeed, most Evaluators $E^{(j)}$ gave scores lower than their respective mean values of $\mathrm{\Phi}$. In contrast, DeepSeek-VL2 gave a relatively high score of 0.940, which was 0.121 points higher than the mean value of $\mathrm{\Phi}$ for DeepSeek-VL2. This suggests that this model favored its own generation.

\cref{fig:qualitive-3} presents an additional example, in which the MLLMs evaluated a caption generated by LLaVA-1.5 in the reference-free setting. Despite a critical error (misidentifying ``rice crispy balls'' as ``marshmallows''), LLaVA-1.5 gave the caption a score of 0.932, which was 0.141 points above the mean value of $\mathrm{\Phi}$. In contrast, most Evaluators $E^{(j)}$ gave scores lower than their respective mean values of $\mathrm{\Phi}$. This discrepancy indicates that LLaVA-1.5 exhibits a self-preference bias.

\subsection{Quantitative Results}
\label{sec:appendix_qualitative}

\begin{figure}[!t]
    \centering
    \includegraphics[width=0.55\linewidth]{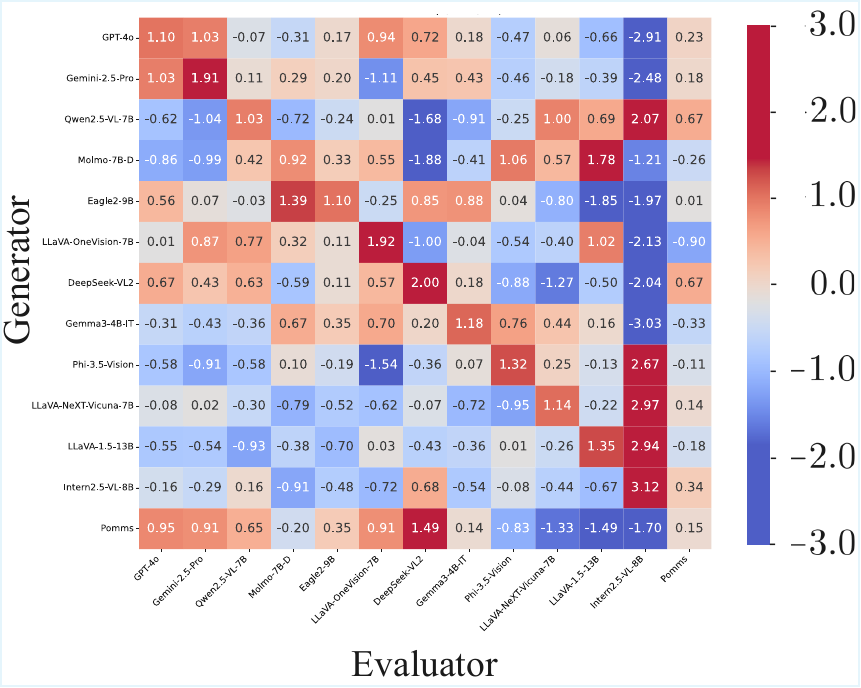}
    \caption{\textbf{Visualization of a full matrix} $\tilde{\mathrm{\Phi} }$ \textbf{reconstructed by incorporating \textsc{Pomms}.} The scores given by \textsc{Pomms} were significantly lower than the corresponding philautia scores of the individual constituent MLLMs.}
    \label{fig:pomms_matrix}
\end{figure}

\cref{fig:pomms_matrix} displays the full matrix $\tilde{\mathrm{\Phi}}$ reconstructed by incorporating \textsc{Pomms}. This instance of \textsc{Pomms} uses the same mixture described in \cref{sec:exp}, consisting of GPT-4o, InternVL2.5-8B, Eagle2-9B, LLaVA-OneVision-7B, and DeepSeek-VL2. The figure illustrates that the scores given by \textsc{Pomms} were significantly lower than the corresponding philautia scores of the individual constituent MLLMs. Furthermore, this reduction in score magnitude was also observed for models that were not part of the \textsc{Pomms} mixture.

Even the most extreme score given by \textsc{Pomms} ($\tilde{\mathrm{\Phi}}_{\text{\textsc{Pomms}}}(\text{LLaVA-OV})=-0.90$) is smaller in magnitude than the smallest philautia score of $0.92$ in Molmo. Because Molmo exhibited the smallest philautia score among all MLLMs in this experiment, this result implies that the scores given by \textsc{Pomms} were smaller in magnitude than the philautia score of any individual model.

\section{Prompts for Generators and Evaluators}
\vspace{-3mm}
\label{sec:appendix_prompt}
We employed the following prompts for the Generators and Evaluators.
For all Generators, we adopted the prompt used in LLaVA-OneVision~\cite{llavaov}: ``Provide a one-sentence caption for the provided image.'' In contrast, for all Evaluators, we employed the prompt from G-VEval \cite{tong2024gveval}, because this is a representative MLLM-as-a-Judge. Although this prompt requests a score on a $0$-$100$ scale, we linearly normalized all evaluation scores to $[0,1]$.
The full prompts for the Evaluators in both the reference-based and reference-free settings are shown below:

\clearpage
\TColorBox{Full prompt for Evaluators in the reference-based setting}{
You will be given one sentence of visual caption generated from one image.\\
\footnotesize
\vspace*{1\baselineskip}
Your task is to rate the generated caption on one metric.\\
\vspace*{1\baselineskip}
Please make sure you read and understand these reference captions carefully. Please keep these references open while reviewing, and refer to them as needed.\\
\vspace*{1\baselineskip}
Evaluation Criteria:\\
\vspace*{1\baselineskip}
Score is from 0 to 100 - selection of important content from the references and the image. The generated caption should accurately describe the important aspects of the image while including the essential information from the references. Annotators were instructed to penalize captions which contained redundancies and excess information.\\
\vspace*{1\baselineskip}
Evaluation Steps:\\
\vspace*{1\baselineskip}
1. Carefully observe the provided image to understand its main content.\\
2. Read the reference captions carefully to identify the important information they highlight.\\
3. Compare the generated caption to both the reference captions and the visual content of the image.\\
4. Assess how well the generated caption covers the main points of the visual content and the reference captions, and how much irrelevant or redundant information it contains.\\
5. Assign an integer score from 0 to 100, considering both the alignment with the image and the inclusion of key points from the references. Please remember the score.\\
\vspace*{1\baselineskip}
Reference captions:\\
\{\{Reference\}\}\\
\vspace*{1\baselineskip}
Image is attached\\
\vspace*{1\baselineskip}
Generated captions:\\
\{\{Caption\}\}\\
\vspace*{1\baselineskip}
Response Format:\\
\vspace*{1\baselineskip}
You should first give a detailed reason for your score, ending with a sentence like this:\\
The final score is \$\{\{score\}\}\$.\\
\vspace*{1\baselineskip}
Note that the score should be an integer from 0 to 100, and should be wrapped in dollar signs (\$).
}
\clearpage
\TColorBox{Full prompt for Evaluators in the reference-free setting}{
You will be given one sentence of visual caption generated from one image.\\
\footnotesize
\vspace*{1\baselineskip}
Your task is to rate the generated caption on one metric.\\
\vspace*{1\baselineskip}
Evaluation Criteria:\\
\vspace*{1\baselineskip}
Score is from 0 to 100 - selection of important content from the image. The generated caption should accurately describe the important aspects of the image. Annotators were instructed to penalize captions which contained redundancies and excess information.\\
\vspace*{1\baselineskip}
Evaluation Steps:\\
\vspace*{1\baselineskip}
1. Carefully observe the image provided.\\
2. Identify the main points of the visual content in the image.\\
3. Assess how well the generated caption covers the main points of the visual content, and how much irrelevant or redundant information it contains.\\
4. Assign an integer score from 0 to 100, please remember it.\\
\vspace*{1\baselineskip}
Generated captions:\\
\{\{Caption\}\}\\
\vspace*{1\baselineskip}
Response Format:\\
\vspace*{1\baselineskip}
You should first give detailed reason for your score, and ending with sentence like this:\\
The final score is \$\{\{score\}\}\$.\\
\vspace*{1\baselineskip}
Note that the score should be an integer from 0 to 100, and should be wrapped in the dollar signs (\$).\\
}

\end{document}